# Enhanced Real-Time 6-DOF Extended Reality Catheter Tracking for Evaluating Potential Improvement in Efficiency, Precision, and Depth Perception for Cardiac Interventions

Mohsen Annabestani[1], Sandhya Sriram[2,3], Andrew Kuzemczak[1], S. Chiu Wong[4], Alexandros Sigaras[2,3] and Bobak Mosadegh[1*]

1. Dalio Institute of Cardiovascular Imaging, Department of Radiology, Weill Cornell Medicine, NY, USA
2. Englander Institute for Precision Medicine, Department of Systems and Computational Biomedicine, Weill Cornell Medicine, NY, USA
3. AI-XR Lab, Department of Systems and Computational Biomedicine, Weill Cornell Medicine, NY, USA
4. Division of Cardiology, Department of Medicine, Weill Cornell Medicine, NY, USA

**Abstract:** Despite advances in 3D ultrasound, most percutaneous cardiac interventions still rely on 2D visualization, limiting depth perception and spatial understanding. To address this challenge, we developed an Extended Reality (XR)-based platform that enables real-time six-degree-of-freedom (6-DOF) catheter tracking and visualization within a patient-specific 3D heart model. The system combines a custom machine-vision algorithm for 5-DOF catheter tracking with a 3D-printed electromechanical encoder that measures catheter roll, providing complete 6-DOF motion reconstruction. In a proof-of-concept study, 20 novice medical students navigated an intracardiac echocardiography (ICE) catheter to six anatomical targets using either immersive 3D visualization or a conventional 2D cathlab-style view. Participants in the 3D condition completed the task in 54.6 seconds and traveled 1,939 mm on average, compared with 267.5 seconds and 7,854 mm in the 2D condition. Therefore, the XR-based 3D system was more than 5× faster and required ~5× less catheter travel. The 3D mode also improved targeting precision and reduced performance variability. Participants consistently rated immersive visualization higher for accuracy, speed, usability, and clinical value. Kinematic analysis showed smoother depth-axis navigation in 3D, whereas 2D users relied on repeated corrective movements. These findings demonstrate that XR-based visualization can substantially improve procedural training efficiency, precision, and motor control.

## 1. INTRODUCTION

Extended Reality (XR) technology, which includes Virtual Reality (VR), Augmented Reality (AR), and Mixed Reality (MR), has been widely promoted as a major step forward for the future of medicine. XR has been gradually integrated into healthcare because it offers important advantages such as scalability, lower costs, realistic simulation capabilities, and the ability to support remote care. In early medical training, several meta-analyses have shown that XR based tools can lead to greater improvements in clinical skills and knowledge acquisition when compared to non-immersive setups[1]. In surgical practice, XR systems have been linked to better procedural accuracy, enhanced technical skills, and reductions in operating room time and procedural complications[2-4]. Beyond surgery, XR has demonstrated potential in mental health care[5], medical education[6, 7], rehabilitation[8, 9], and even in helping cancer patients manage anxiety and stress[10]. Despite these successes, most applications had concentrated on training, preparation, or postoperative care, while the use of XR for clinical procedures remains less frequent. Furthermore, prior work has highlighted the potential benefits of XR across different procedural phases, relatively few efforts have directly tackled real time implementation[11]. This gap is particularly useful in transcatheter interventions, where precise catheter navigation depends on accurate, immediate visualization and interaction by the operators.

Successful catheter-based navigation requires clinicians to guide devices through complex vascular pathways and cardiac chambers with high accuracy to optimized procedural outcome. Traditionally, interventional cardiologists rely on imaging modalities like fluoroscopy and echocardiography, which often provide limited information about the catheter's three-dimensional orientation and position, forcing operators to mentally reconstruct its trajectory during in the procedure[12]. As patient anatomies and pathologies become more complex, the need for improved visualization and more advanced guidance systems continues to grow, particularly systems capable of full six degrees of freedom (6 DOF) tracking and simultaneously accounting for both cardiac cycle and breathing related motion would potentially help increase procedural efficacy[13].

The idea of 6 DOF describes the ability of an object to move through three-dimensional space. It includes three translational directions, which are up or down, left or right, and forward or backward, and three rotational directions, known as pitch, yaw, and roll[13, 14]. In cardiac interventions, complete 6 DOF tracking is essential for precise control, especially for adjunct imaging like intracardiac echocardiography (ICE) catheters and implanting intracardiac medical devices. Although all six degrees are relevant, the roll angle has particular importance because it reflects rotation along the catheter's longitudinal axis. Roll adjustment enables operators to steer the catheter around structures inside the heart, access desired regions more efficiently, and optimize imaging orientation. This rotational control also allows clinicians to accommodate patient specific variations and can shorten procedure times by facilitating rapid adjustments to the imaging field[15, 16]. Procedures like transcatheter valvular implantations and clip placements depend heavily on accurate roll angle control for optimal device deployments[17-20].

Despite its significance, existing clinical imaging systems often do not provide accurate real time roll angle information. These limitations affect both clinical workflows and training environments. To address this challenge, we developed a 3D printed electromechanical encoder capable of real time roll angle measurement for ICE catheters, along with an advanced XR based platform for full 6 DOF catheter tracking and visualization. Using XR headsets, operators can view real-time digital information superimposed onto their natural environment, enabling an intuitive presentation of 2D and 3D guidance content. These capabilities are valuable for catheter-based training[16], pre procedural planning[15-17, 21], and application for enhanced intraoperative guidance [22, 23]. The system offers several advantages, including continuous feedback on catheter position and orientation, accurate roll angle tracking, and an immersive interface that enhances situational awareness. It also addresses the known limitations of 2D fluoroscopic imaging by improving depth perception and spatial understanding. For learners, this approach provides a safer and more cost-effective alternative to fluoroscopy-based training[24, 25], while enabling realistic psychomotor skill development by simulating actual device manipulation within a controlled XR environment[26-28].

The contributions of this work extend beyond transcatheter cardiac procedures and offer broader implications for digital health and medical education. By examining how 3D visualization influences performance in XR based environments, this study contributes to the ongoing shift toward more intelligent and interactive simulation as well as enhanced procedural guidance platforms. Specifically, this work introduces a novel approach for accurate visualization of a catheter's roll angle, a parameter that is often overlooked yet essential for advanced navigation in XR-guided cardiac catheterization. Using the Meta Quest 3 headset and real time 3D catheter reconstruction that reflects the user's physical manipulation, the system produces a realistic catheter model embedded within a patient specific heart representation. This integration results in a more immersive and informative experience compared with conventional techniques. The multi angle visualization tools built into the system can help reduce the learning curve for new catheter technologies and better prepare clinicians for challenging anatomical cases, enhance procedure efficacy and procedural outcomes.

## 2. METHODOLOGY

We developed a novel real-time 6-DOF catheter tracking system which is fully compatible with the Meta Quest 3 XR headset. The system can work with a wide range of commercially available catheters. It consists of two primary hardware components: a 3D-printed vision box and an electromechanical encoder. The 3D-printed vision box consists of two orthogonally positioned cameras that capture biplane images of the catheter (**Figure 1-b, c**). A custom computer vision algorithm analyzes these images to determine the catheter's orientation and shape to reconstruct its 3D configuration (**Figure 1- d, e, f, g**) and extracting real-time 5-DOF features. The electromechanical encoder (**Figure 1-h**), in conjunction with an Arduino board (**Figure 1-i**) and a custom Python program (**Figure 1-j**), enables real-time measurement of the roll angle, thereby completing the 6-DOF tracking.

Real-time catheter data, including the roll angle, is transmitted to the XR headset using the WebSocket protocol (**Figure 1-k**). Within this setup, the Unity game engine drives an independent rendering environment (**Figure 1-n**). This environment displays a reconstructed catheter, and visualized roll angle, within a patient-specific heart model (**Figure 1-m**), providing users with a highly immersive and interactive experience. The heart model is generated from a cardiac CT scan captured in DICOM format at end-diastole (**Figure 1-l**). The catheter data is co-registered with this anatomical representation within the Meta Quest 3 (**Figure 1-o**), allowing users (**Figure 1-p**) to manipulate a real commercial catheter and observe its real-time movement within the 3D heart model,

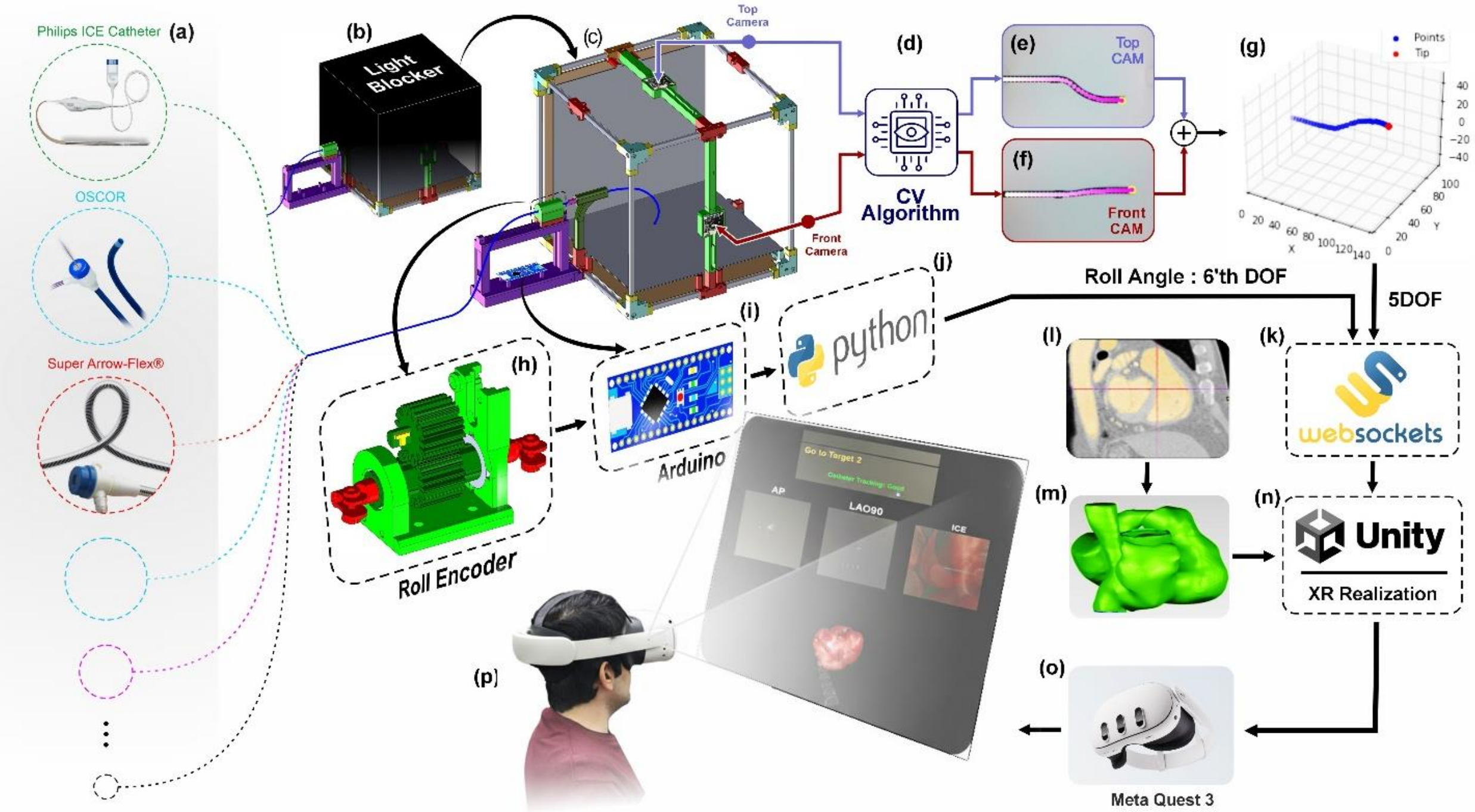

**Figure 1:** The proposed XR-based system: (a) commercially available catheters utilized in the system; (b) vision box with a light blocker; (c) vision box without a light blocker; (d) custom computer vision algorithm; (e & f) inferred catheter shapes derived from biplane views; (g) the reconstructed 3D shape of the catheter; (h) the encoder designed for measuring the roll angle of ICE catheters; (i) an Arduino board for recording and transmitting encoder values; (j) a Python program that converts encoder values into roll angles; (k) a WebSocket unit; (l) segmented blood volume from a de-identified CT scan (DICOM); (m) a patient-specific 3D heart mesh model; (n) a Unity-powered rendering environment; (o) a Meta Quest 3 XR headset; and (p) a user manipulating a real catheter within the 3D model of a patient's heart while observing the process through the headset.

## 2.1. 3D-Printed Vision Box

To enable catheter tracking and provide a stable platform for controlled manipulation, we developed a cubic 3D model using Dassault Systems SolidWorks 2022 and fabricated it with a FDM 3D printer. As shown in **Figure 2**, several key elements were incorporated into the setup to ensure accurate computer vision tracking. A removable inlet, positioned at the origin (0,0,0) of the 3D Cartesian coordinate system, allows catheter insertion and movement within the cube's central open space. This inlet is interchangeable to accommodate catheters of different diameters. Additionally, mounts were integrated into the design to securely hold two cameras in orthogonal positions along two sides of the cube, forming a biplane imaging system. These cameras capture video of the catheter as it moves through the central region of interest (ROI).

To define the ROI, we positioned eight fiducial markers (mTi and mFi for i=1,2,3, and 4, where T and F stand for Top and Front) at specific points on two cross-shaped pillars within the model. These markers ensure that the ROI is appropriately sized targets for a human heart. The cross-shaped pillars, used only temporarily for marker detection, are removable once the markers have been identified. The fiducial markers are placed on orthogonal planes, allowing for the transformation of each camera's viewpoint into a global coordinate system. This coordinate transformation is vital for reconstructing the catheter's 3D trajectory and ensuring proper alignment between the heart model and the catheter inlet (where the catheter is inserted into the Vision Box). To eliminate external light interference and fully control lighting conditions, we enclosed the vision box with light-blocking layers and incorporated internal LED lighting systems, coupled with a PMMA light diffuser for background illumination. The LED planes, laser-cut from translucent PMMA sheets, provide a high-contrast background to enhance the visibility of the catheter for the cameras, regardless of its color or type. This design helps the computer vision algorithm to accurately isolate and segment the catheter. Except for the connecting pillars (as shown in **Figure 2**), the entire setup was 3D printed from PLA thermoplastic using a PRUSA MK4 FDM 3D printer. The components are modular and connected using neodymium magnets for easy assembly and disassembly.

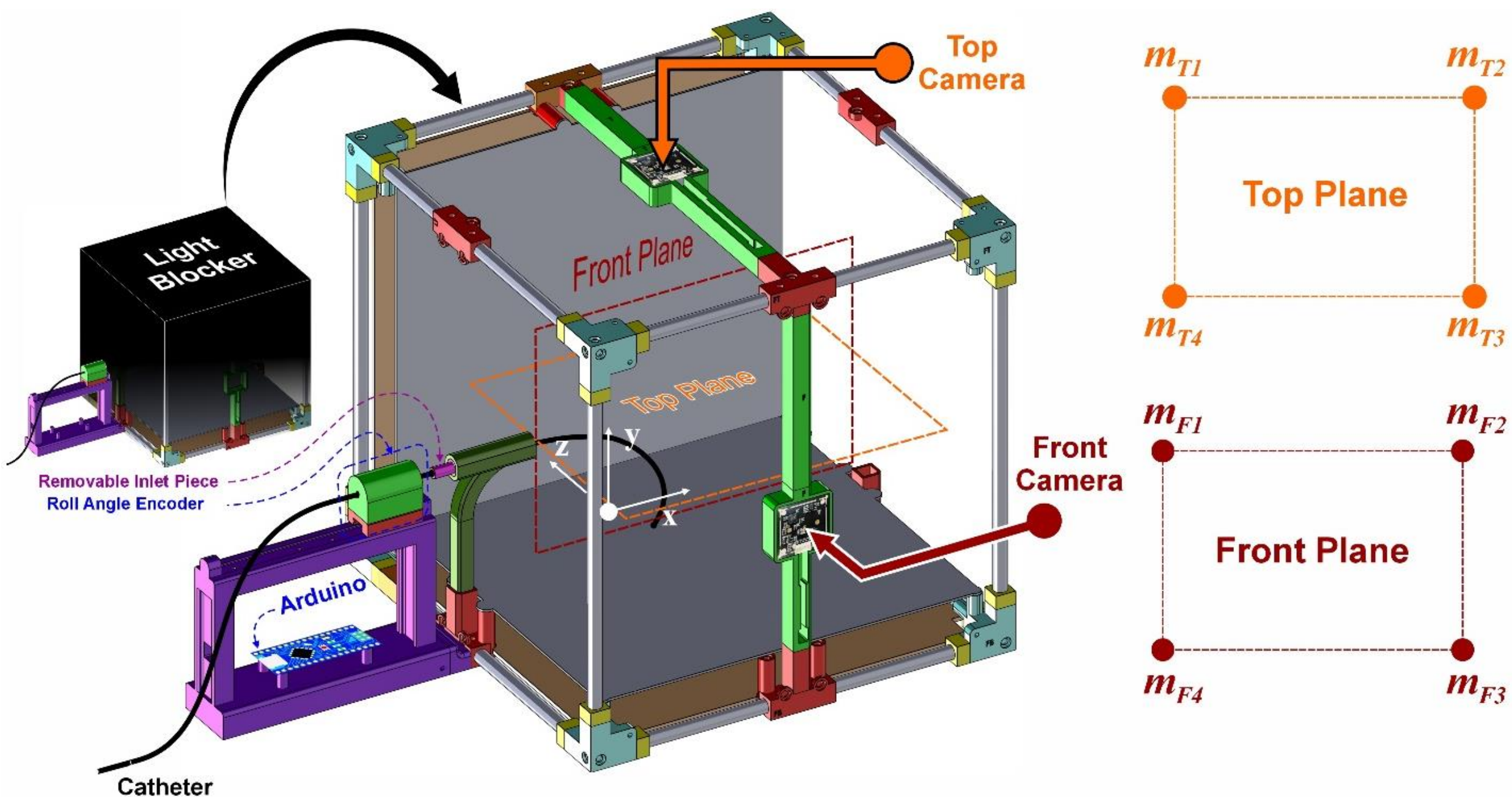

**Figure 2:** The schematic design of the proposed Vision Box includes a removable inlet fixture for catheter insertion at the center of the region of interest (ROI) and a mount for the Roll Angle encoder positioned on a linear rail. The box features four color-coded reference points (mTi and mFi, where i=1,2,3, and 4, with T and F denoting Top and Front), which define the cubic boundary of the heart. Additionally, two mounts are placed on opposite sides of the cube to securely hold two cameras, creating an orthogonal biplane imaging system.

## 2.2. Machine Vision-Based Algorithm for 5DOF Catheter Tracking

We designed a robust and modular computer vision algorithm for 5DOF catheter tracking to address the challenges of real-time procedural guidance with adaptability to diverse imaging modalities, including fluoroscopy (**Figure. 3**). Building on our prior framework[29], the algorithm incorporates a multi-stage pipeline that begins with "*perspective transformation*", rectifying distortions in orthogonal camera feeds by mapping fiducial markers to a unified 3D coordinate system. Each frame, denoted as $I$ ($r$, $c$, $t$), where $r$, c, and $t$ represent pixel row, column, and time dimensions, undergoes *"pre-processing"* to enhance catheter visibility. This includes Gaussian low-pass filtering (LPF) to suppress high-frequency noise and adaptive contrast/brightness adjustments to mitigate uneven illumination—common artifacts.

Central to the algorithm is the *"segmentation"* module, which isolates the catheter from the background using adaptive thresholding with dynamic local per-pixel thresholds. This is the only module requiring modification when transitioning from camera-based imaging to fluoroscopy, as fluoroscopy introduces distinct noise profiles (e.g., X-ray scatter, lower contrast, and uneven background) that necessitate specific segmentation module. Raw segmentation often introduces discontinuities (gaps) or false positives (artifacts), such as glare from reflective catheter surfaces or residual noise from shadows in camera-based images or overlapping anatomical structures in fluoroscopy images. To address this, two critical submodules are employed: *"artifact removal"* and *"gap filling"*. The artifact removal submodule leverages morphological closing operations (e.g., dilation followed by erosion) to eliminate small, disconnected regions, combined with connected-component analysis to retain only the largest contiguous structure—the catheter. The gap filling submodule reconstructs discontinuities using cubic spline interpolation, guided by curvature continuity constraints to preserve the catheter's natural bending profile. These submodules are indispensable for ensuring a contiguous, artifact-free segmentation mask, as gaps or extraneous structures would otherwise propagate errors into subsequent skeletonization and trajectory reconstruction steps.

Following segmentation refinement, the *"skeletonization"* module generates a 1-pixel-wide medial axis representation of the catheter using Zhang-Suen thinning [30], followed by *"pruning"* to remove residual spurious branches caused by minor segmentation irregularities. The skeletonized catheter is then processed by the *"directional feature extraction"* module, where a Breadth-First Search (BFS) algorithm [31] traces the catheter's trajectory from tip to tail. The BFS implementation begins by initializing a queue with the catheter's distal tip (identified via intensity gradient analysis) and iteratively explores all neighboring skeleton pixels at the current depth before progressing to subsequent layers. This ensures the shortest path is prioritized, minimizing deviations caused by loops or bifurcations. To optimize efficiency, the algorithm discards backtracking paths and enforces a directional bias toward the catheter's proximal end using a heuristic based on historical curvature. Skeleton pixels are treated as nodes in a graph, and BFS iteratively explores neighboring pixels at increasing depths, prioritizing the most continuous path to avoid dead-ends. This approach is particularly effective in navigating complex curvatures or bifurcations that traditional depth-first search (DFS) methods might misinterpret.

The orthogonal 2D trajectories from the top-view and front-view cameras are fused in the *"post-processing"* module. Pixel coordinates are scaled to real-world millimeters using pre-calibrated camera intrinsics, and the 3D trajectory is reconstructed via triangulation. To balance computational efficiency with anatomical fidelity, the trajectory is down sampled to $K$ uniformly spaced control points, with the distal tip and entry location fixed as the first and last points, respectively. Intermediate points are dynamically interpolated to capture bending profiles without overfitting to local noise.

Finally, the "*live data streaming*" module serializes the 3D coordinates of all K points into a JSON payload, enabling real-time integration with XR systems such as the Meta Quest 3. This modular architecture ensures flexibility: transitioning from camera-based imaging to fluoroscopy requires only modifications to the segmentation module, while all other components remain unchanged. Implemented in Python with OpenCV and custom libraries, the computer vision pipeline operates at 24 FPS for real-time processing, exceeding the frame rate of standard fluoroscopy systems commonly used in clinical catheterization laboratories (15 FPS). The pipeline achieves precise 3D pose reconstruction and sub-millimeter tip localization accuracy (<1 mm average error), as validated in prior work [29], enabling dynamic visualization of catheter movement within patient-specific anatomies.

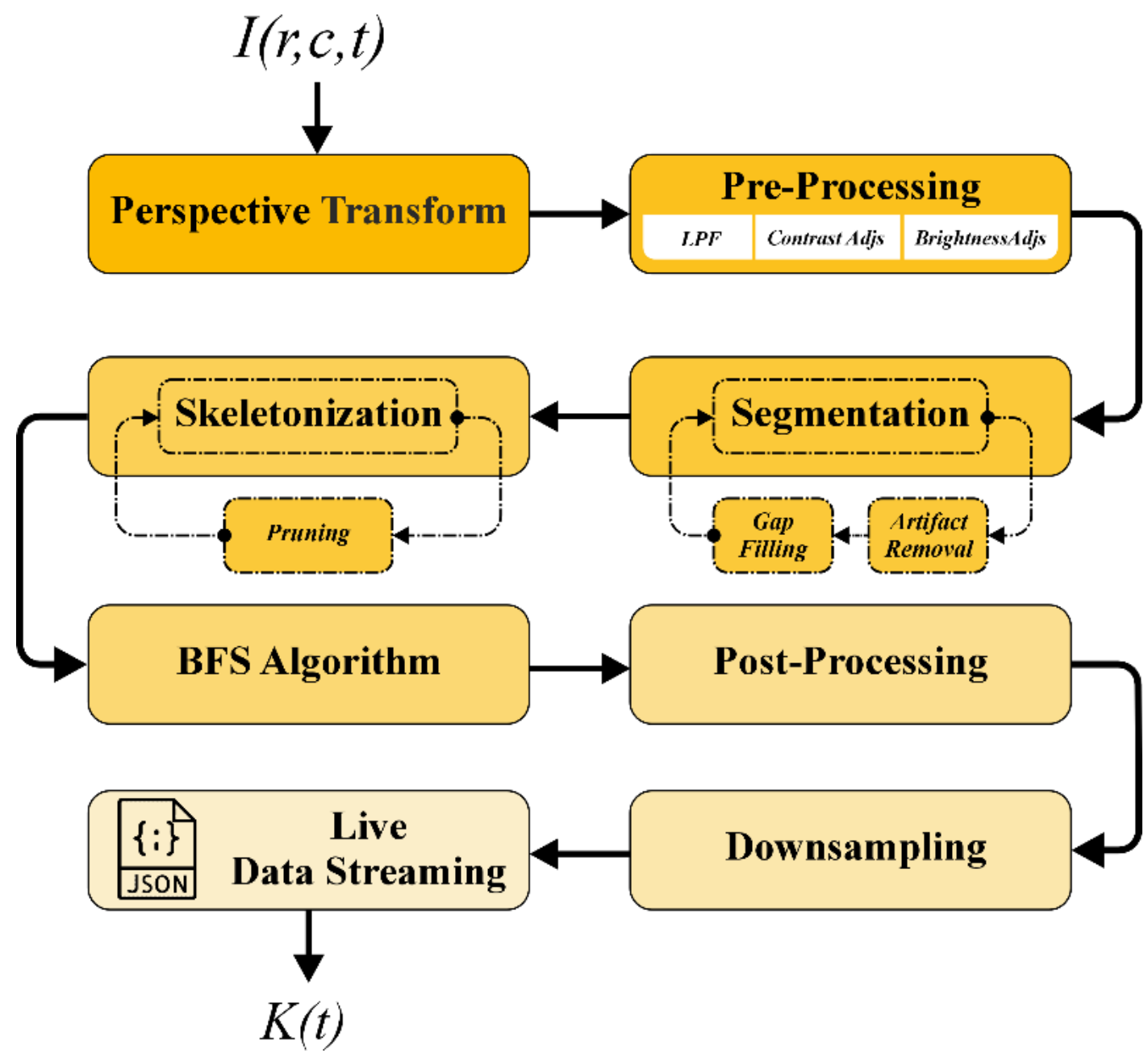

**Figure 3:** Block diagram of proposed machine vision algorithm.

## 2.3. Roll Angle Encoder for Required 6$^{th}$ DOF of ICE Catheter

We designed and implemented an electromechanical encoder system to measure the roll angle of an ICE catheter, as illustrated in **Figure 4**. The system comprises a gearbox, including two gears, where the bottom gear securely holds the ICE catheter using two soft TPU locks, ensuring a firm grip without damaging the catheter. When the catheter rotates, the bottom gear transfers this rotation to the top gear, which is directly connected to a rotary encoder. The rotary encoder, in turn, sends real-time signals to an Arduino Nano, which communicates with a Python script to calculate the catheter's roll angle in real time.

The entire setup, including the gearbox and rotary encoder, is mounted on a base that slides along a linear rail, ensuring that the ICE catheter's locking mechanism does not interfere with its linear movement into the vision box. All components — the gearbox, encoder, base, and rail — form a modular system that can be easily attached to the vision box when tracking the ICE catheter.

To ensure proper alignment, the system is designed to keep the catheter centered with the vision box inlet. During setup, we designate the right side of the vision box as the zero-degree roll position. A custom jig holds the catheter's transducer securely, ensuring accurate angle calibration. When the gear system is unlocked, the catheter can freely slide along the rail into the gearbox, allowing it to reach the end boundary of the vision box where the transducer jig is located. Once the ICE transducer is placed in the jig, the gear's transducer marker, labeled "T", should be aligned with the right side of the box, after which the catheter should be locked in place. This setup ensures the ICE catheter's zero-degree roll position is properly defined, allowing for smooth movement along the rail to the heart model's boundary.

The Arduino Nano is programmed once using Arduino language, and further processing, including roll angle calculation and automatic COM port detection, occurs within Python. For real-time data acquisition, the system operates with a baud rate of 115200. The system was modeled in SolidWorks and fabricated using a PRUSA MK4 3D printer with PLA thermoplastic material. This modular, easily attachable system ensures precise, real-time roll angle measurement without impeding the catheter’s movement, providing a practical solution for real-time tracking of ICE catheters.

The roll angle encoder was designed and tested for the clinically relevant range of motion for intracardiac echocardiography catheter manipulation. While comprehensive characterization across the complete 0-360° theoretical range was not performed, the encoder operates reliably across the clinical ICE range needed for procedural guidance.

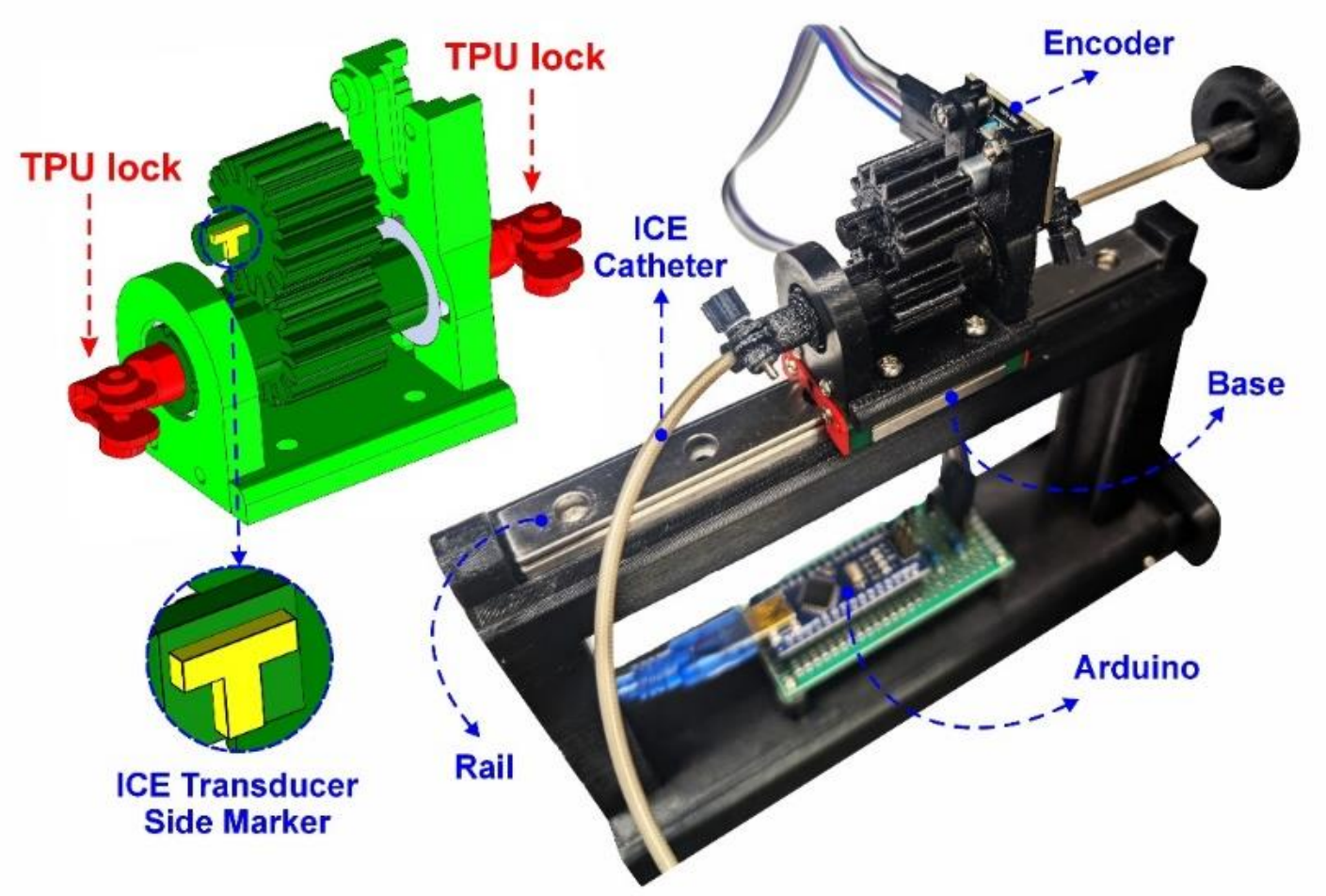

**Figure 4:** Proposed Roll Angle Encoder.

## 2.4. Extended Reality (XR) Rendering

Extended Reality (XR) is employed to integrate real-time catheter tracking with a patient-specific heart model, providing an immersive visualization experience. This XR environment, rendered using the Unity game engine on the Meta Quest 3 headset, combines the spatial configuration of the catheter with an accurate 3D representation of the patient’s heart. The heart model used in this study was generated from a cardiac CT scan acquired at end-diastole and processed using Materialise Mimics and Geomagic Wrap, as described in our previous work [29]. The model remained static throughout all experiments and did not incorporate cardiac cycle dynamics or respiratory motion. This static model approach was appropriate for the current proof-of-concept training study because the laboratory environment was intentionally controlled and simplified, the focus was on demonstrating training benefits rather than clinical implementation, and the system was designed as an educational platform where static anatomical references are suitable. However, real-time clinical deployment would require dynamic cardiac models and continuous registration updates to account for cardiac motion.

The catheter's 3D position, derived from the real-time K-point data (K(t)) output by our computer vision algorithm, is rendered in the XR scene alongside the heart model. The catheter is rendered as a spline object in Unity, with knots on the spline being dynamically added and modified depending on the K-point data to form the shape and curve of the catheter. Communication between the Python-based tracking system and Unity game engine is facilitated via Flask, using a WebSocket connection to ensure real-time data transmission. The user connects the system by entering an IP address and port, allowing dynamic updates to the catheter's position and curvature within the virtual heart.

A crucial feature of the system is the visualization of the ultrasound beam emitted from the tip of the ICE catheter. This beam is represented as a cone-shaped (let's call it **Cone**) field extending from the catheter's tip (**Figure 5**), providing real-time feedback on the roll angle of the catheter. The visualization of the Cone helps trainees and physicians accurately align the ICE catheter transducer, ensuring proper orientation and positioning within the specific target in the heart. This is particularly valuable for catheter guidance, as the roll angle directly influences the direction of the beam, helping users locate the appropriate anatomical structures. Additionally, a spherical marker (let's call it **Dot**) is placed within the scene (at the end of the Cone) as a reference point for reaching the targets, allowing users to practice precise catheter placement and alignment. This visual guidance helps the user to locate the transducer in the optimal position, enhancing both the training and real-life procedural accuracy by providing clear visual feedback on catheter movement and positioning.

For accurate mapping, fiducial markers in the 3D heart model are used as reference points, enabling an affine transformation that aligns the catheter's movements with the heart's anatomical structure. This ensures that the catheter is correctly positioned within the heart model, creating a cohesive and interactive XR experience. The system provides quantitative feedback on catheter movements including distance and angle differences from the target, allowing trainees to practice complex catheterization procedures with real-time guidance, making it a valuable tool for both training simulations and potential clinical applications.

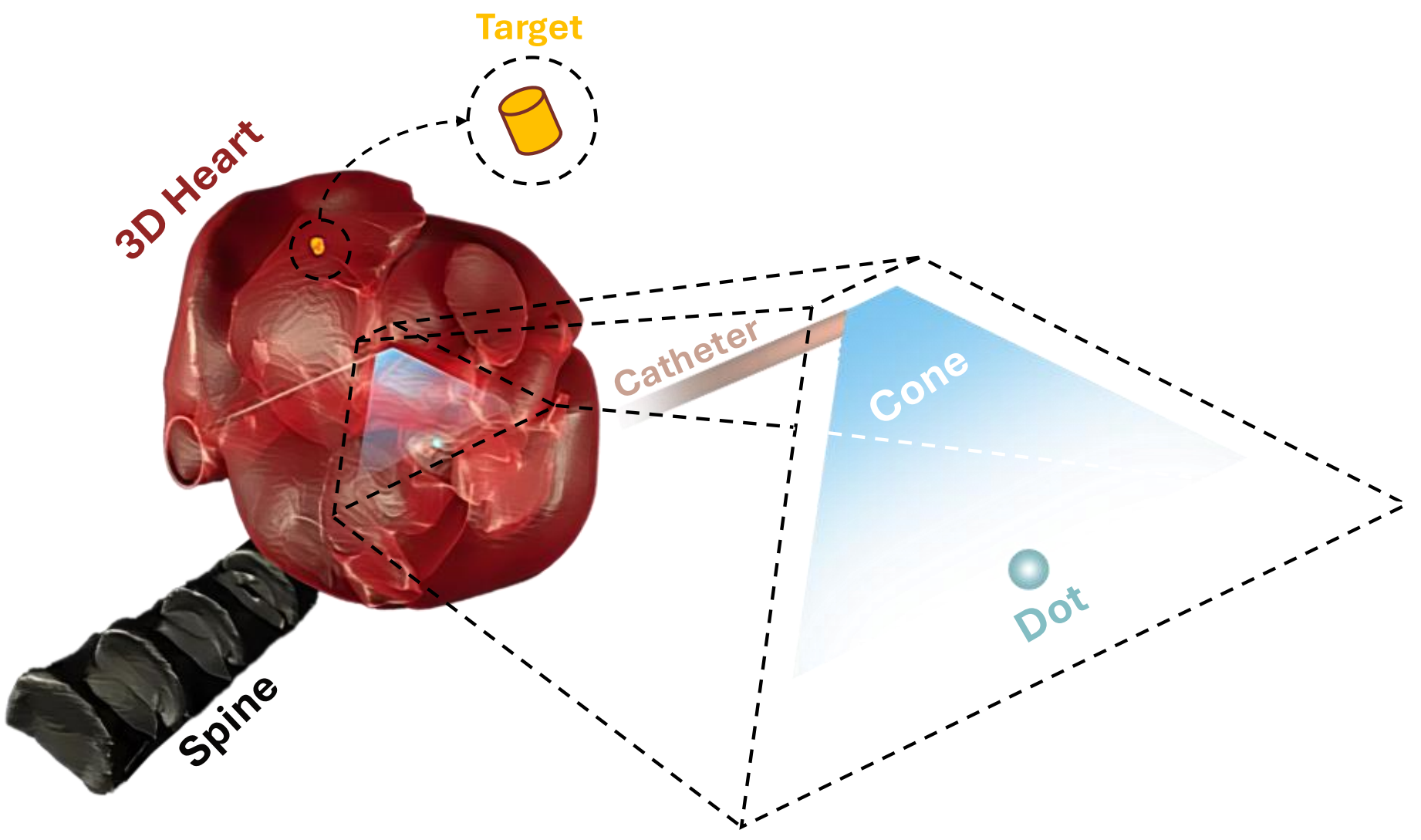

**Figure 5:** The visualization of the ultrasound beam emitted from the tip of the ICE catheter in the 3D rendering.

### 2.5. 2D Mode vs 3D Mode Visualization Study

The aim of this section is to quantitatively and qualitatively evaluate the efficiency, precision, and motor control performance of catheter articulation using our XR-based 6-DOF catheter tracking system, and to compare it with a simulated conventional 2D cathlab–style visualization method. We hypothesize that the added depth perception and spatial awareness offered by our immersive 3D environment will result in measurably better performance.

We recruited 20 second-year medical students from Weill Cornell Medicine (12 men, 8 women; mean age 25.15 ± 2.46 years) to assess the impact of our XR-based image guidance system on operators' ability to reach predesignated targets in a simulated cardiac structure. This cohort was intentionally selected because they represent novice users with foundational knowledge of cardiovascular anatomy but no practical experience with catheter manipulation, providing a standardized baseline for this proof-of-concept training study. Ethical review and approval were waived for this study due to its focus on technology development and non-generalizable conclusions. Participants completed the tasks under two visualization conditions:

1. **3D Mode:** A single immersive, floating 3D patient-specific heart model was displayed with full catheter visualization and tracking in 3D space (**Figure 6-Right**). Participants could freely move their heads, enabling the use of parallax and natural depth cues for distance estimation.
2. **2D Mode:** To simulate visualization approaches commonly used in structural heart interventions, this mode presented two grayscale views, Anteroposterior (AP) and Left Anterior Oblique at 90 degrees (LAO90), without the three-dimensional heart model (**Figure 6-Left**). A third window displayed a simulated intracardiac echocardiography image. This simplified 2D visualization was intentionally designed as a controlled comparison condition and is particularly relevant to structural heart interventions, such as transcatheter aortic valve replacement, left atrial appendage occlusion, and mitral valve interventions, where guidance often relies primarily on echocardiographic imaging rather than extensive fluoroscopic roadmapping. Although this condition does not fully replicate all aspects of clinical fluoroscopy navigation, it provides a controlled framework for comparing conventional 2D projection-based visualization with immersive 3D visualization, which represents the core technical contribution of this work.

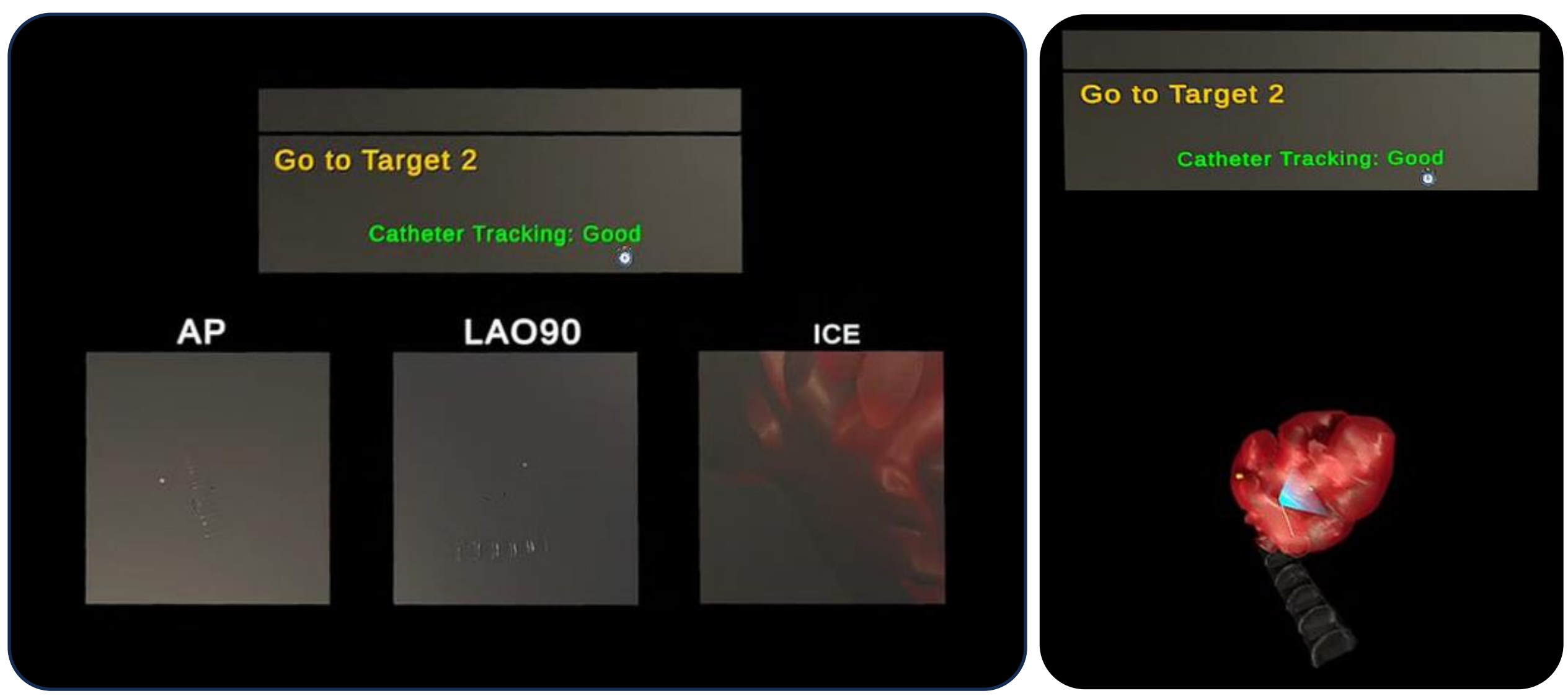

**Figure 6.** Visualization modes. **Right:** Immersive 3D view with full anatomical and catheter visualization. **Left:** Simulated 2D view with AP, LAO90, and ICE projections only.

### 2.6. Onboarding Procedure

To ensure that all participants began the experiment from a consistent baseline, each individual completed a standardized onboarding procedure before starting the main tasks.

1. **Instruction:** Participants were first given an overview of the Vision Box tracking system, the VeriSight Philips ICE catheter, and the relevant patient-specific cardiac anatomy used in the study.
2. **Familiarization:** They then practiced manipulating the physical catheter, including the four-way knob controls (flexion, extension, left, and right) and the roll encoder. During this stage, the instructor demonstrated the recommended way to hold, move, and rotate the catheter inside the Vision Box.
3. **Practice Session:** Participants entered the XR preview environment for an instructor-guided practice session in a combined 2D and 3D setup (**Figure 7**). This unified scene allowed them to view all elements that would appear in the main experiment while giving them the freedom to focus on either mode as needed. They were allowed unlimited practice time and proceeded only once they confirmed that they felt comfortable reaching all predefined targets. This approach helped minimize any performance bias associated with initial unfamiliarity.

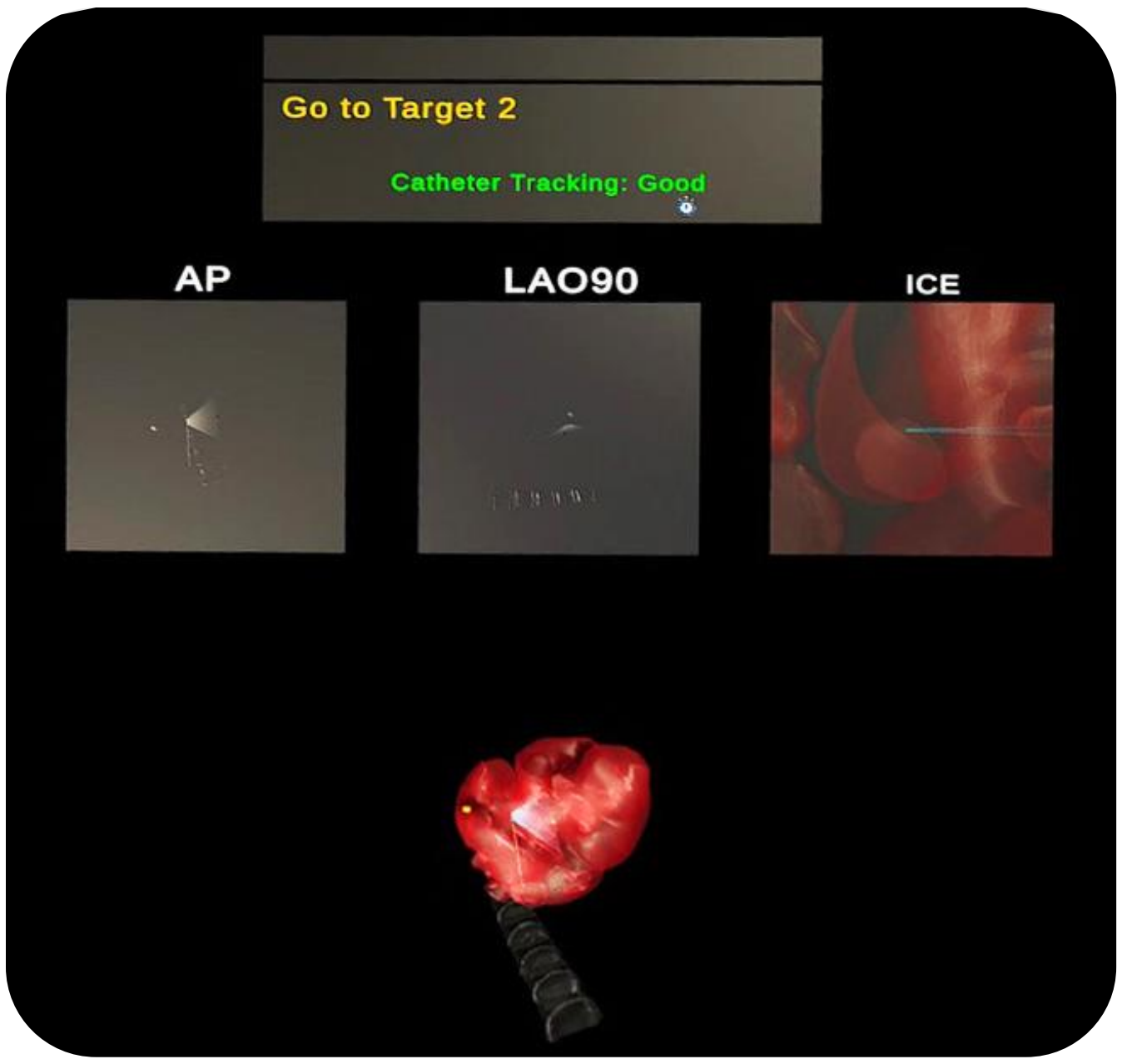

**Figure 7.** Combined 2D and 3D XR practice scene used during onboarding, allowing participants to explore all visual elements before the main experiment.

## 2.7. Experiment Steps

The main experiment required participants to reach a series of six anatomical targets within the heart model in both the 2D and 3D visualization modes. The targets were not selected to represent disease- or procedure-specific landmarks; instead, they were sparsely distributed throughout the three-dimensional heart model to provide diverse navigation challenges across different spatial regions and distances. This design enabled isolation of the effect of visualization modality (3D vs. 2D) on motor control and procedural learning without the confounding influence of complex procedural constraints or specialized anatomical targets.

- **Target Sequence:** The order and identity of the targets were identical for all participants and across both visualization modes to ensure that performance differences were not affected by variations in navigation path length. Targets were considered successfully reached when the catheter tip came within 10 mm of the target center, representing clinically relevant approach accuracy for structural interventional cardiology procedures. The mean distance-to-target values reported in the Results section reflect the average distance maintained during the approach phase before achieving the 10 mm criterion.
- **Session Order:** The 2D and 3D sessions were administered in a randomized and counterbalanced order. An external binary random generator assigned the session sequence for each participant, minimizing potential learning effects and fatigue-related bias. Future clinical validation studies should replace these generic anatomical targets with clinically relevant landmarks specific to structural interventional procedures, such as the aortic valve annulus, left atrial appendage landing zone, or fossa ovalis.

## 2.8. Statistics

Statistical analyses were conducted using the *Shapiro-Wilk test* to assess normality, paired *t-tests* with *Bonferroni correction* for normally distributed measures, and *Wilcoxon signed-rank tests* when normality assumptions were not met. To quantify relative performance between modes, we also computed mode superiority ratios based on subject-level ratios. All analyses were performed on the full cohort of twenty participants, and statistical significance was defined as $p < 0.05$.

# 3. RESULTS

## 3.1. Performance Metrics

We evaluated performance using two primary metrics and two secondary metrics, each capturing complementary aspects of catheter control.

**Primary metrics:**

- **Task Completion Time (*T*)** | The total time in seconds from the start of a trial until the catheter Dot reached all six targets. **Figure 8** shows the results for all subjects.
- **Total Travel Length (*L*)** | The cumulative distance in millimeters traveled by the catheter Dot. This metric was computed as the sum of Euclidean distances between consecutive tracked positions. **Figure 9** shows the results for all subjects.

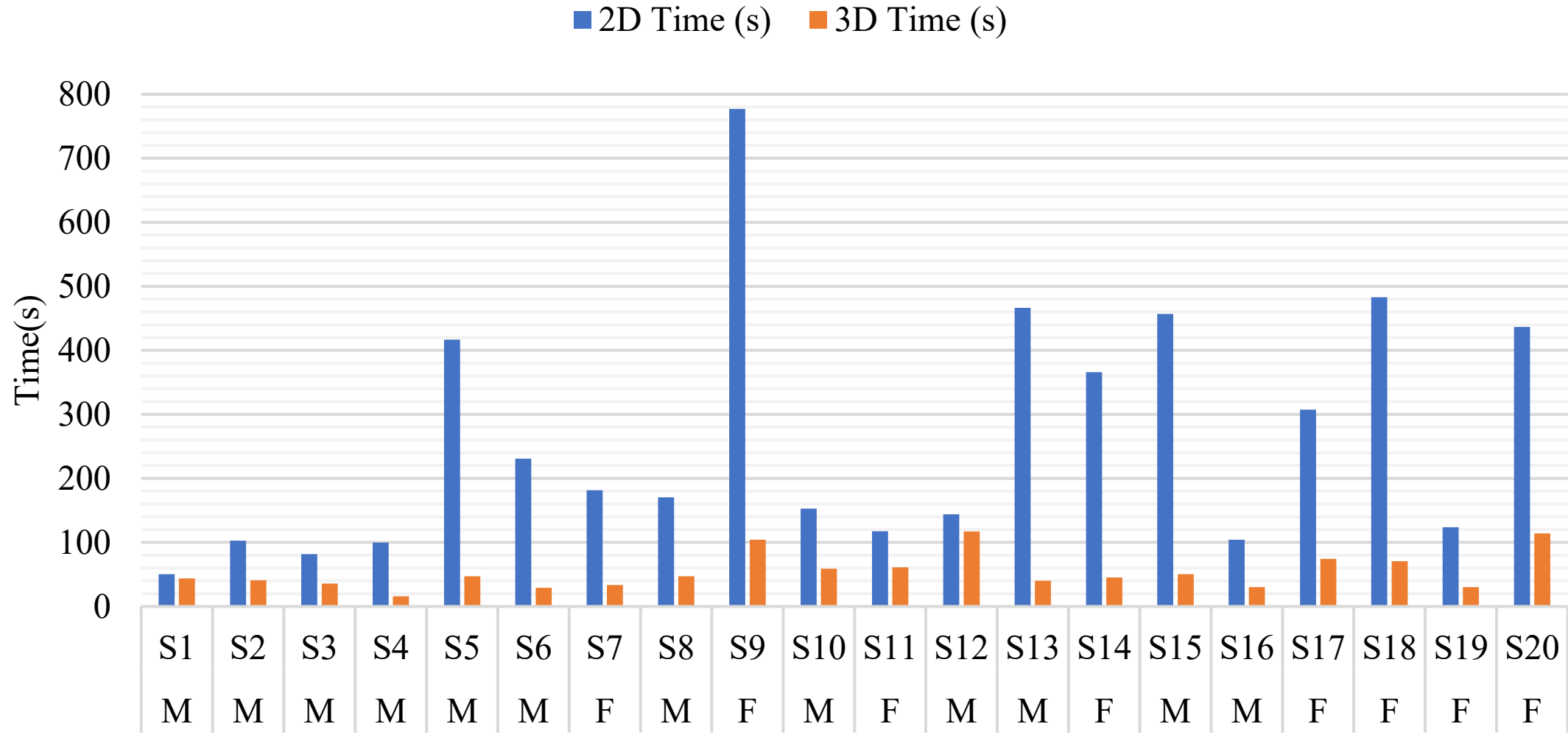

**Figure 8.** Task Completion Time (T) for all participants. Each horizontal pair of bars represents one subject (S1–S20), with sex indicated below each label (M or W). Time reflects the total duration required for the catheter dot to reach all six targets.

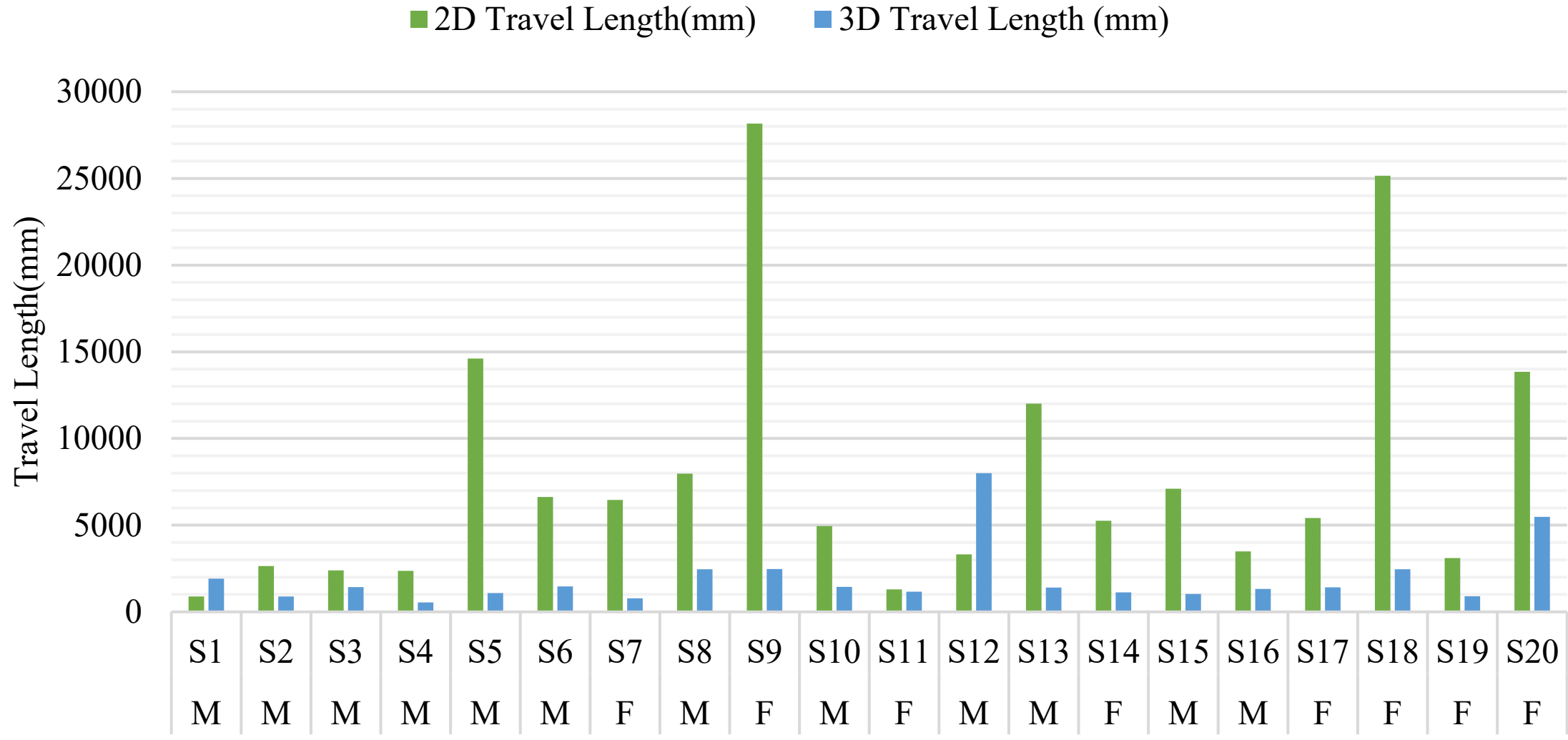

**Figure 9.** Total Travel Length (L) for all participants. Each horizontal pair of bars represents one subject (S1–S20), with sex indicated below each label (M or W). Travel length is the cumulative distance in millimeters traveled by the catheter dot, calculated as the sum of Euclidean distances between consecutive positions.

**Secondary metrics:**

- **Mean Distance to Target ($D_M$)**| The average distance in millimeters between the catheter Dot and the center of the target throughout the active manipulation period.
- **Distance Standard Deviation ($D_{SD}$)** | A measure of movement stability, representing how much the catheter Dot's distance from the target varied over time.

Across all measured outcomes, the immersive 3D mode showed a strong and statistically significant advantage compared with the 2D mode (**Table 1**).

**Table 1.** Paired Comparisons for All Subjects (N = 20)

| Metric | Mean Diff | Test Statistic | p-value (corr) | 2D/3D Ratio | Interpretation |
|---|---|---|---|---|---|
| ***T (s)*** | 208.89 | W = 0 | < 0.001* | 5.16 | 3D is 5.16× faster. |
| ***L (mm)*** | 5914.85 | W = 14 | < 0.001* | 4.91 | 3D requires 4.91× less travel. |
| ***$D_M$ (mm)*** | 5.83 | t = 3.91 | 0.0029* | 1.24 | 3D reaches 1.24× closer to targets. |
| ***$D_{SD}$ (mm)*** | 2.25 | t = 2.81 | 0.033* | 1.25 | 3D shows 1.25× less variability. |

*Note: W indicates results from the Wilcoxon signed-rank test and t indicates results from the paired t-test. * A comparison is considered statistically significant if p < 0.05.*

## 3.2. Efficiency (Time and Travel)

The 3D mode was markedly faster and more efficient than 2D mode.

- **Speed:** Participants completed the task in an average of 54.6 seconds in 3D, compared with 263.5 seconds in 2D, meaning the 3D mode was 5.16× faster (W = 0, $p < 0.001$).
- **Movement Economy:** Total travel length showed a 4.91× less traveling. In 2D, participants traveled more than 7,800 mm on average, reflecting extensive searching and unnecessary motion. In contrast, the 3D mode reduced this to approximately 1,939 mm (W = 14, $p < 0.001$).

This substantial difference (**Figure 10**) demonstrates the cognitive burden of inferring 3D spatial relationships from separate 2D views, a challenge largely eliminated by the immersive 3D environment.

### 3.3. Precision and Stability

Performance at the target also improved in 3D.

- **Accuracy:** The 3D mode enabled users to reach targets more precisely than 2D mode, with a mean improvement of 5.83 mm (2D: 33.5 mm vs 3D: 27.7 mm, t-value = 3.91, p-value = 0.0029).
- **Stability:** Variability in the catheter Dot's distance from the target was lower in 3D comparing to 2D (2D: 13 mm vs 3D: 10.8 mm, t-value = 2.81, p-value = 0.033), indicating that participants could maintain a steadier position once they arrived at the target.

### 3.4. Extreme Performance Gap

The contrast between modes was so large that only one participant (S1) completed the 2D task faster than the group mean time in 3D. Similarly, only two participants (S1 and S11) achieved a travel length in 2D shorter than the group mean in 3D. These observations highlight the considerable inefficiency of the 2D mode for most novice users compare to 3D.

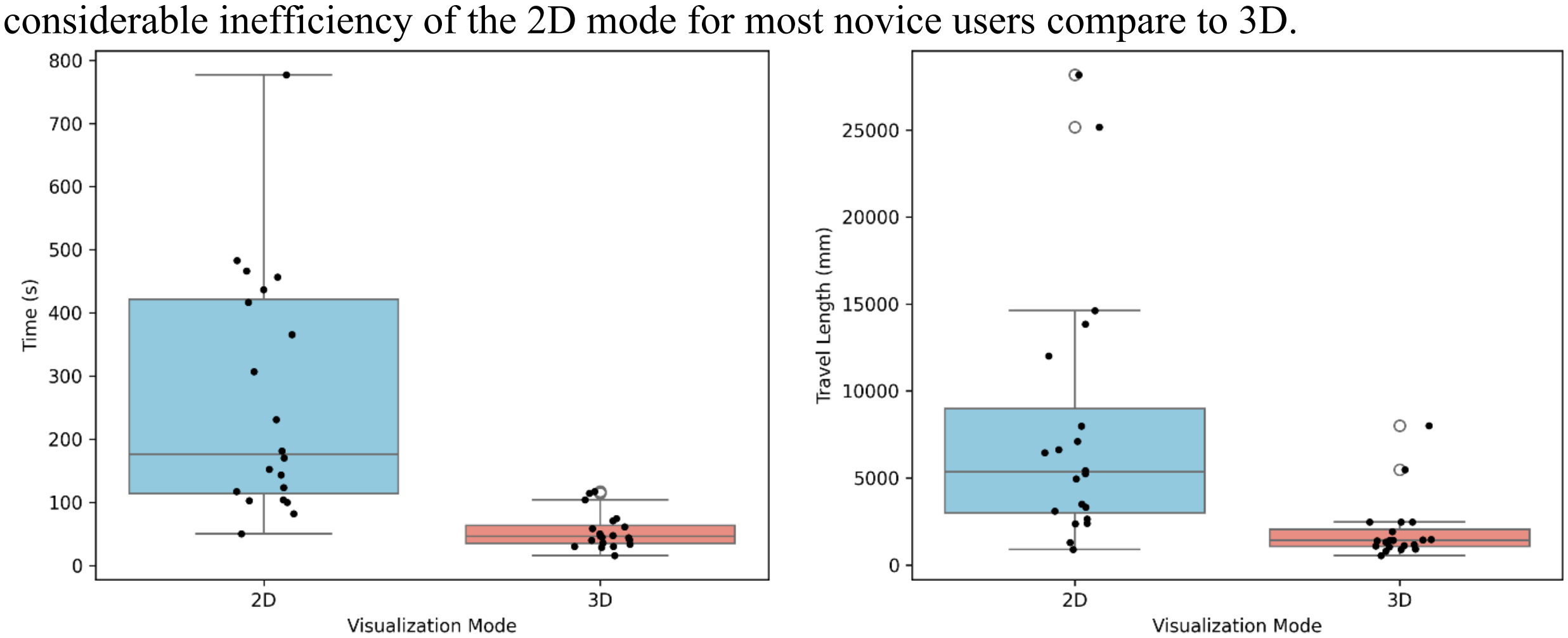

**Figure 10.** Efficiency comparison between 2D and 3D modes. **Left:** Box plot of Task Completion Time (T), showing that 3D is much faster. **Right:** Box plot of Total Travel Length (L), showing that 3D requires much less traveling. The data highlight the efficiency advantage of the immersive 3D mode over 2D.

### 3.5. Post-Hoc Power Analysis and Sample Size Justification

To evaluate the adequacy of the study sample size (n = 20), a post-hoc statistical power analysis was performed using paired-samples t-tests with a significance level of $\alpha = 0.05$ (two-tailed). The analysis demonstrated high achieved statistical power across all primary outcome measures. Task completion time ($T$) showed a Cohen's d effect size of 1.19 with 99.9% statistical power, while total travel length ($L$) demonstrated a Cohen's d of 0.82 with 93.3% power. Mean distance to target ($D_M$) yielded a Cohen's d of 0.90 with 96.7% power, and distance standard deviation ($D_{SD}$) showed a Cohen's d of 0.65 with 78.2% power. The average achieved statistical power across all evaluated metrics was 92.0%.

These findings indicate that the sample size was sufficient to detect the observed differences between the immersive 3D and conventional 2D visualization conditions. Moreover, the consistently large effect sizes across all performance metrics (Cohen's d ranging from 0.65 to 1.19) support the conclusion that the advantages observed with immersive 3D visualization represent robust and practically meaningful improvements in training performance rather than statistical artifacts related to insufficient sample size.

### 3.6. Qualitative Analysis and Subject Reports

After completing the trials, participants provided feedback through a qualitative survey using a 0-5 Likert scale, supported by written reports. Overall ratings were highly positive for the XR system's immersive features and clinical utility (**Table 2**). The 3D visualization was considered extremely helpful, with a mean score of 4.75 (SD = 0.44) for accuracy and 4.70 (SD = 0.47) for speed, and 95% and 90% of participants selecting 4 or 5, respectively. In contrast, the 2D "ICE View" received a mean score of only 1.75 (SD = 1.12), with 70% of participants rating it 2 or below, highlighting the challenges of spatial interpretation and depth perception in the 2D mode. Participants reported that the system interface was user-friendly (mean = 4.35, SD = 0.67), and they perceived strong clinical value in the XR system. Ratings for pre-operative training and planning averaged 4.85 (SD = 0.37), and intra-operative procedures averaged 4.90 (SD = 0.31), with 75% of responses giving a perfect score of 5. Overall experience was rated excellent, with a mean of 4.85 (SD = 0.37) and 90% of participants selecting 5. These results closely align with the objective performance improvements in speed, precision, and stability observed in section 3.3.

**Table 2**: Subjective Survey Ratings (Likert Scale 0-5)

| Survey Item | Mean Rating | SD |
|---|---|---|
| 3D Helpfulness for Accuracy (vs 2D) | 4.75 | 0.44 |
| 3D Helpfulness for Speed (vs 2D) | 4.70 | 0.47 |
| ICE View Helpfulness in 2D | 1.75 | 1.12 |
| Usefulness for Pre-Op Training | 4.85 | 0.37 |
| Usefulness for Intra-Op Procedures | 4.90 | 0.31 |
| Interface User-Friendliness | 4.35 | 0.67 |
| Overall System Experience | 4.85 | 0.37 |

## 4. DISCUSSION

Our current study clearly demonstrates that immersive 3D visualization provides substantial advantages over traditional 2D displays for intracardiac catheter navigation. These benefits extend across multiple metrics, including efficiency (time and travel), precision, and stability. We hypothesize that one of the key reasons for this superiority is the lack of depth perception in the 2D mode and the enhanced motion control afforded by the 3D mode. In this section, we discuss these factors in detail. We also examine how prior video game experience (VGE) influences participant performance. We hypothesize that high VGE has a stronger impact on 2D performance, as most video games are non-immersive experiences like the 2D mode. In contrast, the fully immersive nature of the 3D mode may act as a performance standardizer, reducing performance differences associated with VGE.

### 4.1. Kinematic Motion Analysis and Depth Perception

To gain a clearer understanding of how depth perception impacts efficient catheter manipulation, we analyzed the 3D trajectory of the catheter movement (x, y, z) over time. Because the catheter is inserted primarily along the X-axis, which represents depth (**Figure 11-Left**), we expected depth-related limitations in the 2D mode to leave a measurable footprint in the movement pattern. The Y and Z axes represent lateral and vertical adjustments, which are typically distinct from forward advancement. By examining how users distributed their motion across these axes, we aimed to determine whether the absence of depth cues in 2D selectively altered movement along the depth dimension or produced a broader change in the strategy of navigation.

As an initial metric, we compared the total distance traveled along each axis in 2D and 3D. If total distance were a strong marker of depth-related difficulty, the ratio of 2D over 3D should be largest along the X-axis. Instead, the ratios were nearly identical across all axes: 3.98 for X, 4.14 for Y, and 4.03 for Z. As shown in **Figure 11-Right**, although total travel in 2D is clearly longer than in 3D, this metric does not reveal a selective deficit along the depth direction. This reinforces that total distance alone is not a sensitive footprint of depth perception. Rather, the more informative indicators of depth-related difficulty arise from the structure and behavior of the motion itself, which is why the subsequent Principal Component Analysis (PCA), and velocity analyses are essential for capturing how operators compensate for missing depth cues and how their movement strategies fundamentally differ between 2D and 3D environments.

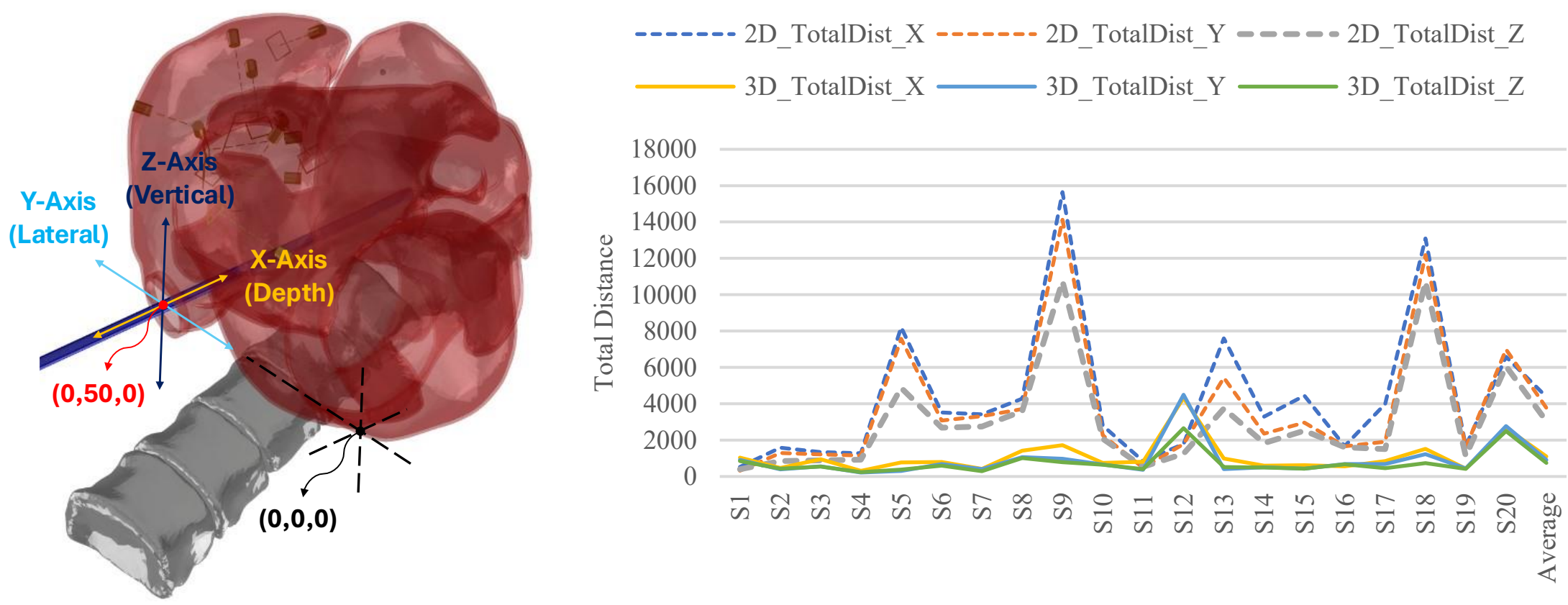

**Figure 11. Left:** Coordinate system showing X as the insertion (depth) axis. **Right:** Total travel in 2D versus 3D across axes.

#### 4.1.1. Principal Component Structure (PCA) and Depth Perception

We applied PCA to the 3D trajectory matrix to identify the dominant directions of catheter motion. PCA is a statistical method that reduces complex movement data into a small number of meaningful components. In this context, PCA reveals the main directions in which the catheter moves and helps us understand how users plan their path in space. The first principal component (PC1) captures the direction with the greatest variability, essentially reflecting the primary behavioral pattern during navigation.

Across both visualization modes, approximately half of the movement variance was explained by PC1 (2D: 0.545; 3D: 0.544), which is expected given the physical constraint that catheter insertion predominantly occurs along the X (depth) axis. What differentiates the modes is how this primary movement is constructed. PCA loadings (**Figure 12**) quantify the relative contribution of each spatial axis (X, Y, Z) to PC1. In 2D mode, loadings were strongly skewed toward the Y (0.695) and Z (0.544) axes, with only a minor contribution from X (0.179). This imbalance suggests that participants struggled to perceive and control movement along the depth direction (X axis). Instead of making deliberate forward or backward adjustments, they compensated by relying heavily on lateral and vertical motions, reflecting the absence of reliable depth cues. Together, these patterns clearly show a *lack of depth perception in 2D*.

By contrast, in 3D mode, the loadings (Figure 12) were more evenly distributed across all three axes (X: 0.374, Y: 0.566, Z: 0.570). Although the lateral and vertical axes still contributed, the substantially higher weighting on the X axis, which is more than twice that of 2D, indicates improved alignment between the intended insertion direction and the actual movement. Instead of searching through lateral or vertical adjustments, participants in 3D mode produced a more coherent and direct trajectory that matched the catheter's true spatial path more closely. Taken together, these results show that *depth perception in 3D was significantly improved*, with compensation along the X axis more than two times higher compared to 2D.

When considered alongside the axis-specific distance results, the PCA analysis shows that the improvement seen in 3D mode is not only due to reduced total movement but also to a meaningful shift in the organization of motion. In the 3D environment, participants were able to identify and follow the correct depth direction earlier and with greater certainty, which minimized unnecessary lateral and vertical corrections and resulted in a more direct and efficient navigation pattern. These combined observations clearly demonstrate a *lack of depth perception in 2D and a significantly improved depth perception in 3D*.

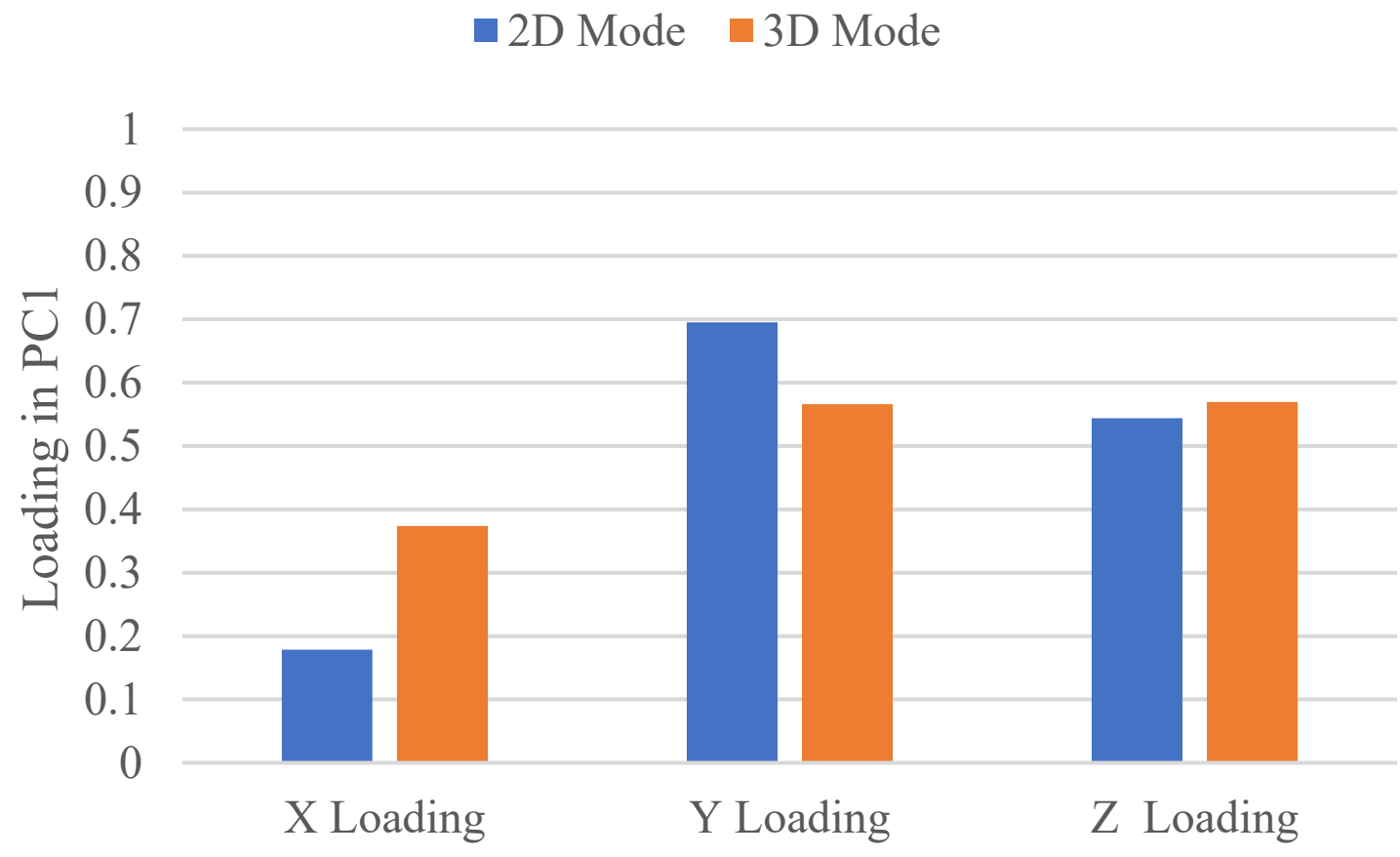

**Figure 12.** PCA loadings for PC1 showing axis contributions to catheter movement. In 2D, motion is dominated by Y/Z (0.695/0.544) with minimal depth (X: 0.179). In 3D, loadings are more balanced (X: 0.374, Y: 0.566, Z: 0.570), reflecting improved depth-aligned navigation.

### 4.1.2. Mean Velocity, Confidence, and Depth Perception

In kinematic terms, mean velocity is defined as

$$v_{mean} = \frac{\Delta x}{\Delta t}$$

which depends on net displacement ($\Delta x = x_1 - x_0$) rather than total distance traveled. This means a participant who moves in a straighter, more consistent direction will have a higher mean velocity, even if the total path length is shorter. Conversely, frequent reversals, pauses, or back-and-forth adjustments reduce net displacement over time and therefore lower the mean velocity.

Analysis of the mean velocities of our data supports this interpretation and shows again the lack of depth perception. Along the depth or the same insertion axis of catheter (X), participants in 3D mode achieved a mean velocity of 0.104, which is almost three times higher than the 0.039 observed in 2D, despite covering less total distance in 3D. This pattern aligns with more continuous forward progression in 3D and more interrupted, hesitant motion in 2D. In 2D, the low X-axis mean velocity is consistent with repeated small corrections that cancel each other out over time, reducing the net forward displacement.

Overall, the mean velocity results align with the broader finding that 3D visualization reduces uncertainty along the depth axis, enabling more efficient forward motion, while 2D visualization leads to lower net X-axis progress due to limited depth perception.

## 4.2. Prior Experience and Skill Equalization

This study examined how prior experience, specifically self-reported Video Game Experience (VGE), influenced performance. VGE is utilized as a proxy for 2D screen-based spatial skills, reflecting an individual's familiarity with abstract spatial manipulation on screens, which can strongly affect performance in tasks requiring the interpretation of 2D projections. For this analysis, participants were grouped based on their VGE score (0-5): the Low VGE Group consists of scores less than 3 (VGE $< 3$), and the High VGE Group consists of scores greater than or equal to 3 (VGE $\geq 3$). By comparing these two groups, we determined the extent to which prior experience influenced task execution and how immersive visualization mitigated this effect.

### 4.2.1. VGE and the 2D Skill Gap

In the 2D mode, a substantial disparity in performance was observed that was highly associated with the VGE level. Participants were segmented into the High VGE Group and the Low VGE Group. The High VGE Group consistently demonstrated significantly faster task completion and more efficient catheter navigation compared to the Low VGE Group. For instance, the ratio of 1.85 in 2D Time (Low VGE / High VGE) shows the High VGE Group was 85% faster at completing the task, while the ratio of 2.52 in 2D Travel shows the High VGE Group exhibited superior efficiency by traveling more than 2.5 times less in total traveling. This large performance gap suggests that individuals in the Low VGE Group experienced a higher cognitive load in the 2D mode, requiring them to mentally reconstruct 3D relationships from multiple 2D views, which resulted in slower, more hesitant movements.

However, Immersive 3D visualization effectively acted as a performance standardizer, significantly reducing this outcome disparity driven by VGE differences. The immersive environment allowed all participants to rely on intuitive hand-eye coordination and natural depth perception, which fundamentally reduced the performance advantage previously conferred by high VGE. In the 3D mode, the efficiency gap in Travel (mm) was nearly eliminated, with the ratio dropping sharply to 0.92 (Low VGE / High VGE). The gap in Time (s) also narrowed considerably, with the ratio dropping to 1.28. These results (**Table 3**) demonstrate that the 3D visualization environment particularly benefits participants with lower baseline 2D skills (Low VGE Group), promoting uniform training efficacy and procedural performance across different experience levels.

**Table 3:** Performance Comparison and Skill Equalization by VGE Level

| Metric | Mode | High VGE Group (Mean) | Low VGE Group (Mean) | Ratio (Low VGE / High VGE) | Interpretation |
|---|---|---|---|---|---|
| **Time (s)** | 2D | 187.74 | 347.22 | 1.85 | High VGE Group ≈85% faster |
| **Time (s)** | 3D | 47.96 | 61.21 | 1.28 | Time Gap Narrows |
| **Travel (mm)** | 2D | 4461.89 | 11246.16 | 2.52 | High VGE Group travels more than 2.5 times less |
| **Travel (mm)** | 3D | 2018.35 | 1860.01 | 0.92 | Efficiency Gap Eliminated |

## 4.3. Novel Contributions and Advancements Beyond Prior Work

The present study builds upon our previous work[29], which established the foundational computer vision framework for 5-DOF catheter reconstruction and the patient-specific heart model generation pipeline from cardiac CT imaging. The current work substantially advances this framework by introducing a custom 3D-printed electromechanical encoder for catheter roll-angle measurement, enabling complete real-time 6-DOF catheter tracking within an integrated XR visualization system. In addition, the vision algorithm was further optimized for faster real-time performance, enabling stable operation at 24 FPS for dynamic procedural visualization.

Beyond the hardware and software integration advances, this study provides the first structured comparative evaluation of immersive 3D versus conventional 2D visualization for intracardiac catheter navigation. Through a controlled user study with 20 participants, the system demonstrated significant improvements in procedural efficiency and precision, including approximately 5× faster task completion and markedly reduced travel distance in 3D mode. Furthermore, kinematic analysis revealed how immersive depth perception alters navigation strategies, while gaming-experience analysis showed that XR visualization acts as a performance standardizer across users. Together, these contributions establish a complete XR-based catheter training platform that substantially extends the foundational framework introduced in prior work.

## 4.4. Cost-Effectiveness and Accessibility

The proposed XR system offers a strong cost-effectiveness and accessibility advantage due to its low-cost hardware and scalable design. The primary components include an XR headset (approximately $500) and a custom 3D-printed vision box with electromechanical encoder (approximately $200), both representing one-time capital investments that can be reused across multiple training or procedural sessions with negligible per-use costs.
This minimal cost structure enhances the system's potential for wide adoption across institutions with varying financial resources, particularly when compared to more expensive image-guided platforms such as interventional MRI systems. As a result, the proposed approach provides a scalable and economically feasible solution for both training and potential future clinical support applications.

### 4.5. Limitations

This study has some limitations that should be considered when interpreting the results. The sample size was relatively small (n = 20) and included only second-year medical students from a single institution. While this homogeneous novice cohort is appropriate for isolating the effect of visualization modality and demonstrated strong performance differences between 3D and 2D conditions, it limits generalizability to broader populations. In particular, experienced interventionalists may rely on developed visuospatial strategies that could attenuate the observed advantages of immersive 3D visualization. In addition, recruitment from a single institution may introduce selection bias related to specific educational backgrounds, further limiting external validity.
The study was conducted using a static, patient-specific heart model derived from a single end-diastolic CT scan, without incorporation of cardiac cycle dynamics or respiratory motion. Although this simplification is appropriate for a controlled proof-of-concept training study, it does not reflect the temporal variability and anatomical deformation present in real clinical procedures. Future clinical translation would require dynamic cardiac modeling, real-time registration updates, and integration with live imaging modalities such as fluoroscopy and echocardiography, along with validation in real procedural environments.
The 2D comparator condition was intentionally simplified and does not fully replicate clinical fluoroscopy workflows, which often include contrast injections, road mapping, and additional dynamic imaging strategies. Therefore, the reported performance improvements in the 3D condition should not be directly generalized to all fluoroscopy-guided interventions. Nevertheless, the comparison remains valid for evaluating the core effect of immersive 3D versus simplified 2D projection-based visualization.
End-to-end system accuracy was not independently verified against external tracking systems such as electromagnetic tracking, end-to-end latency was not directly measured, and full-range characterization of the roll-angle encoder (0–360°) was not completed. Similarly, XR-to-heart model registration accuracy was not separately quantified, and real-time drift under dynamic conditions was not assessed.
The anatomical targets were generic and not disease- or procedure-specific, which limits direct clinical interpretability. Likewise, the system was tested only on normal adult anatomy, and its performance in pathological or congenital cases remains untested. The current roll-angle encoder is specifically designed for intracardiac echocardiography catheters and may not generalize to other interventional devices without modification. Collision detection, anatomical constraints, and haptic feedback were not implemented, meaning the simulation does not fully replicate realistic procedural constraints. Finally, the study evaluated the system only as a training platform; clinical application would require substantial further development, including real-time motion compensation, integration with live imaging, clinical validation in experienced operators, and regulatory approval.

## 5. CONCLUSION

This study demonstrates the significant benefits of an XR-based 6-DOF catheter tracking and visualization system as a training platform for intracardiac catheter navigation in novice users. The findings show that immersive 3D visualization substantially improves procedural efficiency, precision, spatial understanding, and motor control in a controlled laboratory environment. Participants using the immersive 3D system completed navigation tasks 5.16 times faster and traveled 4.91 times less total distance compared with the conventional 2D mode ($p < 0.001$ for both), while also achieving improved stability and target precision. Kinematic analysis revealed that enhanced depth perception fundamentally changed navigation behavior, enabling smoother and more confident motion along the primary depth axis while reducing corrective movements. In addition, immersive visualization acted as a strong performance standardizer, reducing disparities associated with prior video game experience and highlighting its value for heterogeneous learner populations.

The proposed system combines a real-time machine vision algorithm for 5-DOF catheter reconstruction, an optimized processing pipeline operating at 24 FPS, and a novel 3D-printed electromechanical encoder enabling complete 6-DOF catheter tracking through roll-angle measurement. Integrated with patient-specific 3D heart models in an immersive XR environment, the platform provides a low-cost, radiation-free, and scalable alternative to conventional fluoroscopy-based training approaches. Unlike interventional cardiac magnetic resonance imaging (iCMR), which provides dynamic real-time anatomical imaging but requires specialized infrastructure, high cost, and device compatibility constraints, the proposed XR framework is designed as a complementary visualization and training platform based on standard pre-procedural imaging commonly used in structural heart interventions. Rather than replacing existing imaging modalities, the system enhances intuitive spatial understanding of catheter position and orientation within patient-specific anatomy and can integrate more readily with existing fluoroscopy and echocardiography workflows. Although the current work represents a controlled training study using static anatomical models and simplified 2D visualization conditions, the results strongly support the potential of immersive XR visualization to enhance procedural training, medical education, and future intraoperative procedural guidance. Future work should focus on dynamic cardiac modeling, integration with live imaging modalities, and validation in experienced interventionalists to support eventual clinical translation in structural heart interventions.

## AUTHOR CONTRIBUTION

Methodology and Machine vision algorithm design, M.A.; Coding and implementation of the algorithm, M.A.; Design and implementation of the 3D setup, M.A.; Design, Fabricate, program and implementation of the Roll Encoder, M.A.; Testing and investigation, M.A.; Experimental study, Data and Statistical analysis, M.A.; Subject Hiring, A.K.; Subject training and test, M.A. and A.K.; Front-end development and coding, S.S.; Unity implementation; Resources, B.M. and S.C.W.; Front-end project supervision, A.S.; Conceptualization, M.A. and B.M.; Project administration, B.M.; Supervision, B.M.; Funding acquisition, B.M. All authors have read and agreed to the published version of the manuscript.

**Acknowledgement**

We thank Philips for providing the VeriSight ICE catheter through a research contract ("no specific grant number").

**Data Availability**

The data that support the findings of this study are available on request from the corresponding author, BM.

**Competing Interests**

The authors declare no competing interests.

**Ethics Approval Declaration**

All participants were employees at Weill Cornell Medicine, but were not authors of this paper, and provided informed consent to participate in the study, following all relevant guidelines. The Weill Cornell Medicine Institutional Review Board (WCM-IRB) determined this study to be non-human subjects research and was exempted. This decision was based on the fact that no clinical procedures, treatments, or invasive actions were conducted and instead these studies were performed to assess the effectiveness of 2D vs. 3D visualization but was not done in a manner that provides a generalizable conclusion.

**Funding Declaration**

The authors declare that this research received no external funding or financial support.